\documentclass{article}

\PassOptionsToPackage{round,compress}{natbib}

\usepackage[final]{nips_2018}
\usepackage[utf8]{inputenc} 
\usepackage[T1]{fontenc}    
\usepackage{hyperref}       
\usepackage{url}            
\usepackage{booktabs}       
\usepackage{amsfonts}       
\usepackage{nicefrac}       
\usepackage{microtype}      
\usepackage{multirow}
\usepackage{graphicx}
\usepackage{amsmath}
\usepackage{amssymb}
\usepackage{soul}
\usepackage{color}
\usepackage{xspace}
\usepackage[title]{appendix}

\usepackage[caption=false]{subfig}
\usepackage{svg}

\definecolor{LinkColor}{rgb}{0.0, 0.18, 0.39}
\hypersetup{colorlinks=true,
    linkcolor=LinkColor,
    citecolor=LinkColor,
    filecolor=LinkColor,
    menucolor=LinkColor,
    urlcolor=LinkColor}

\title{Toward Scalable Neural Dialogue State Tracking Model}

\author{
Elnaz~Nouri\\
  Microsoft Research and AI\\
  \texttt{elnouri@microsoft.com} \\
  \And{Ehsan~Hosseini-Asl}\\
  Salesforce Research\\
  \texttt{ehosseiniasl@salesforce.com} 
}

\begin{document}

\maketitle

\begin{abstract}

The latency in the current neural based dialogue state tracking models prohibits them from being used efficiently for deployment in production systems, albeit their highly accurate performance. 
This paper proposes a new scalable and accurate neural dialogue state tracking model, based on the recently proposed Global-Local Self-Attention encoder (GLAD) model by~\cite{Zhong2018GlobalLocallySD} which uses global modules to share parameters between
estimators for different types (called slots) of dialogue states, and uses local modules to learn slot-specific features.
By using only one recurrent networks with global conditioning, compared to (1 + \# slots) recurrent networks with global and local conditioning used in the GLAD model, our proposed model reduces the latency in training and inference times by $35\%$ on average, while preserving performance of belief state tracking, by $97.38\%$ on turn request and $88.51\%$ on joint goal and accuracy. Evaluation on Multi-domain dataset (Multi-WoZ) also demonstrates that our model outperforms GLAD on turn inform and joint goal accuracy. 

\end{abstract}

\maketitle

\section{Introduction}

Dialog State Tracking (DST) is an important component of task-oriented dialogue systems. DST keeps track of the interaction's goal and what has happened in the dialog history. Majority of the deployed dialogue systems in commercial settings such as common customer support systems and intelligent assistants, such as Amazon Alexa, Apple Siri and Google Assistant, take advantage of dialogue state tracking. Dialog state tracking uses the information from user utterance at each turn, context from previous turns, and other external information as well as the output of the system at every turn. 
Decision made by the dialogue state tracker, is then used to determine what action should be taken by the system in next steps. This is a critical role to play in the design of any task oriented dialogue system.  


State of the art approaches for dialogue state tracking rely on deep learning models,
which represent the dialogue state as a distribution over all candidate slot values that are defined in the ontology. 
Recently, several neural-based DST systems have been proposed.~\cite{Mrksic2017NeuralBT} proposed a Neural Belief Tracker (NBT) model based on binary decision making of each state-values, where representation of user utterance, system action, and candidate pairs are computed based on deep distribiutional representation of word vectors. In their model, they used deep network (DNN) and convolutional network (CNN) to compute such representation vectors.~\cite{Wen2017ANE} proposed a sequence-to-sequence model for estimating the next dialogue state. In their work, the encoded hidden vector of user utterance is used to determine the current dialogue state, followed by a policy network to query over knowledge databse. Then, the retrieved information is used as a conditionining input to the decoder, to generate the system response.

Recently,~\cite{Zhong2018GlobalLocallySD} proposed a model based on training a binary classifier for each slot-value, Global-Locally Self Attentive encoder (GLAD, by employing recurrent and self attention for each utterance and previous system actions, and measuring similaity of these computed representation to each slot-value, which achieve state of the art results on WoZ~\citep{Wen2017ANE} and DSTC2~\citep{Williams2013TheDS} datasets.

Although the proposed neural based models achieves state of the art results on several benchmark, they are still inefficient for deployment in production system, due to their latency which stems from using recurrent networks. 
 In this paper, we propose a new encoder, by improving GLAD architecture~\citep{Zhong2018GlobalLocallySD}. The proposed encoder is based on removing slot-dependent recurrent network for utterance and system action encoder, and employing a global conditioning of aforementioned encoder on the slot type embedding vector. By removing the slot-dependent recurrent network, the proposed model is able to preserve the performance in predicting correct belief state, while improving computational complexity. The detailed description of encoder is explained in the section~\ref{sec:proposed}.

\subsection{Related Works}
\label{seq:related}
A similar scalable dialogue state tracking model is also proposed by~\cite{Rastogi2017ScalableMD}, which is based on conditioning the encoder input. They used a similar conditioning of user utterance representation on slot values (candidate sets) and slot type. 
However, our proposed model is based on conditioning only on slot type. 
Therefor, our proposed model is simpler since it contains only one conditioned encoder for user utterance, whereas ~\cite{Rastogi2017ScalableMD} model requires two independet conditioned encoder. 

Recently,~\cite{Xu2018AnEA} proposed a model for unknown slot type by using a pointer network, based on conditioning to slot type embedding. Our proposed model is also relaxing the current GLAD architecture for unknown slot types during inference.

\begin{figure*}[t]
\centering
\subfloat[GCE Encoder]{
\hspace{-1em}\centering\includegraphics[width=0.55\textwidth]{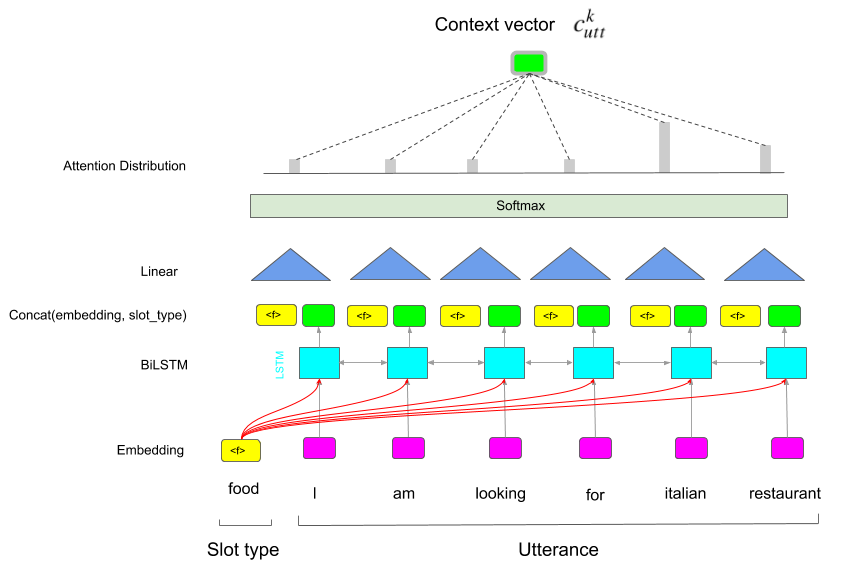}}\hspace{1em}
\subfloat[Dialogue State Tracker ]{\includegraphics[width=0.4\textwidth]{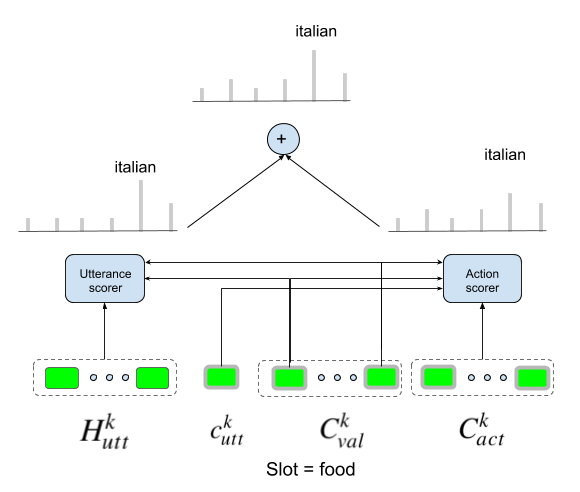}}
  
  \caption{Proposed Dialogue State Tracking model with (a) Globally-Conditioned Encoder (GCE), and (b) overall state tracking model. }
  \label{fig:gce-encoder}
  \end{figure*}

\section{Proposed model}
\label{sec:proposed}

In this section, we describe the proposed model. First, section~\ref{seq:glad} explains the recently proposed GLAD encoder~\citep{Zhong2018GlobalLocallySD} architecture, followed by our proposed encoder in section~\ref{seq:proposed}.

\subsection{Global-Locally Self-Attentive Model}
\label{seq:glad}
GLAD model is based on learning multiple binary classifier for each slot-value pair. In this architecture, separate encoders are considered for utterance, previous system action, and all slot values. The output of these encoders are then used by two scores model, i.e. previous system action and utterance, to predict the probability distribution on slot-value pairs. This means, each scores model compute the similarity of each slot-value to the utterance representation or previous system action. All encoders used similar architecture, i.e. global-local self attention (GLAD). To compute the hidden representation of its input sequence and its summary (context), GLAD a combination of bidirectional LSTM~\citep{Hochreiter1997LongSM}, to compute the temporal representation, followed by self-attention layer to extract the context vector. To incorporate information regarding each slot, there is a dedicated recurrent and self-attention network for each slot. Therefor, to estimate the probability distribution over values of each slot, GLAD encoder has to learn a different hidden and context vector for utterance and previous system action.  


\subsection{Globally-Conditioned Encoder (GCE)}
\label{seq:proposed}
In this section, we describe the proposed globally-conditioned encoder (GCE) model. Here, we employ the similar approach of learning slot-specific temporal and context representation of user utterance and system actions, as proposed in GLAD~\citep{Zhong2018GlobalLocallySD}. However, we emphasize the limitation of GLAD encoder in using slot-specific recurrent and self-attention layers in their encoders. Our proposed encoder is based on improving the latency and speed of inference by remving the inefficient recurrent layers and self-attention layers, without degrading the performance. 

The proposed model is based on removing slot-specific recurrent and self-attention layers, and using only slot embedding vector (i.e. $s_{k}$ for $k$-th slot), as a conditioning vector to the temporal and context extraction layers, as shown in Figure~\ref{fig:gce-encoder}.   

\begin{align}
    H^{k} & =  \text{biLSTM}(f(X,s_{k}))\in \mathbb{R}^{n\times d_{rnn}} \\
    a_{i}^{k} & =  W f(H^{k}_{i}, s_{k}) + b \in \mathbb{R} \\
    p^{k} & =  \text{softmax}(a^{k}) \in \mathbb{R}^{n} \\
    c^{k} & =  \sum_{i}{p^{k}_{i} H^{k}_{i}} \in \mathbb{R}^{d_{rnn}}
\end{align}

To compute $k$-th slot-based representation $H^{k}$, the slot embedding $s_{k}$ is concatenated with sequence tokens $X$, i.e. user utterance or previous system actions, as input to the recurrent layer, where concatenation denoted as $f(X, s_{k})$. 
Then a slot-based attention score $a^{k}_{i}$ is computed for each token hidden representation $H^{k}_{i}$, by concatenating them to slot embedding $s_{k}$ and passing to a linear layer. In this way, the computed attention is conditioned on the slot embedding, to pay attention to the slot-only information in the input sequence $X$.

Therefor, the GCE encoder function can be represented as,
\begin{align}
    \text{encode} : X, s_{k} \rightarrow H^{k}, c^{k} 
\end{align}

\paragraph{Encoding Modules:} Based on the definition of the proposed GCE encoder, the representation of user utterance, previous system action and current slot-value pair is computed as below, 

\begin{align}
    H^{k}_{utt}, c^{k}_{utt} = \text{encode}(U,s_{k}) \\
    H^{k}_{act_{j}}, C^{k}_{act_{j}} = \text{encode}(A_{j}, s_{k}) \\
    H^{k}_{val}, c^{k}_{val} = \text{encode}(V, s_{k})
\end{align}

where $U$ denotes the user utterance word embeddings, $A_{j}$ is the $j$-th previous system action, and $V$ is the current slot value pair to be evaluated (e.g \textit{food=italian}).

\paragraph{Scoring Model:} We follow the proposed architecture in GLAD~\citep{Zhong2018GlobalLocallySD} for computing score of each slot-value pair, in the user utterance and previous system actions.

To determine whether the user has mentioned a specific value of slot $k$, we compute the slot-$k$th conditioned scores for its values.
\begin{align}
    a^{k}_{utt_{i}} & = (H^{k}_{utt_{i}})^\intercal c_{val}^{k} \in \mathbb{R} \\
    p_{utt}^{k} & = \text{softmax}(a_{utt}^{k}) \in \mathbb{R}^{m} \\
    q_{utt}^{k} & = \sum_{i}{p^{k}_{utt_{i}} H^{k}_{utt_{i}}} \in \mathbb{R}^{d_{rnn}} \\
    y^{k}_{utt} & = W q^{k}_{utt} + b \in \mathbb{R}
\end{align}

Similarly, to determine whether any slot-value is mentioned in previous system actions, that the user is referring to in the current utterance, we compute slot-conditioned scores of previous $j$ system actions. 

\begin{align}
    a^{k}_{act_{j}} & = (C^{k}_{act_{j}})^\intercal c^{k}_{utt} \in \mathbb{R} \\
    p^{k}_{act} &= \text{softmax}(a^{k}_{act}) \in \mathbb{R}^{l+1} \\
    q^{k}_{act} &= \sum_{j}{p^{k}_{act_{j}} C^{k}_{act_{j}}} \in \mathbb{R}^{d_{rnn}} \\
    y^{k}_{act} &= (q^{k}_{act})^\intercal c^{k}_{val} \in \mathbb{R}
\end{align}

The final scores of slot $k$ is the weighted sum of user-based and action-based scores, i.e. $y^{k}_{utt}$ and $y^{k}_{act}$, which are normalized by sigmoid function $\sigma$.
\begin{align}
    y &= \sigma (y^{k}_{utt} + \omega y^{k}_{act}) \in\mathbb{R}
\end{align}
where $\omega$ is a learned parameter.

\section{Experiment}
In this section, we evaluate the proposed encoder for the task of dialogue state tracking om single and multi-domain dialogue state tracking. Wizard of oz (WoZ) restaurant reservation dataset~\cite{Wen2017ANE} is chosen for single-domain, and the performance is compared with the recent neural belief tracking models. Moreover, we also evaluate on recen;t proposed multi-domain dataset, Multi-WoZ~\citep{Budzianowski2018MultiWOZA}, which consists of seven domains, i.e. restaurant, hotel, train, attraction, hospital, taxi, and police. 

The evaluation metric is based on joint goal and turn-level request and joint goal tracking accuracy. The joint goal is the accumulation of turn goals as described in~\cite{Zhong2018GlobalLocallySD}. The fixed pretrained GLoVe embedding~\citep{pennington2014glove} with character-n gram embedding~\citep{Hashimoto2017AJM} are used in embedding layer. The implementation details and code of the GCE model can be found at~\href{url}{https://github.com/elnaaz/GCE-Model}. 

\paragraph{Single-Domain:} Table~\ref{tab:woz} shows the evaluation performance on WoZ dataset. It is indicated that our proposed GCE model performance is on par with GLAD model. To further compare the latency of GCE and GLAD during training and testing, computation time for a batch of turn and the overall epoch time during training is measured. We further evaluate the complete test time, which contains $400$ dialogue and $1646$ turns (WoZ test set), as shown in Table~\ref{tab:time}. The computation time is measured in second, and it is indicated that GCE improves latency in both training and testing by $35\%$ on average. 

\paragraph{Multi-Domain:} Table~\ref{tab:mwoz} shows the evauation on Multi-Woz~\citep{Budzianowski2018MultiWOZA} dataset which consists of $10$k dialogues. In this setting, we completely ignore the domain information and use slot names only. The results indicate that GCE model outperforms GLAD on turn inform and join goal accuracy.

\renewcommand{\tabcolsep}{0.04cm}
\begin{table}[h!]
\caption{Test accuracy on WoZ restaurant reservation dataset.}
\label{tab:woz}
\centering
\begin{tabular}{lcc}
\toprule
 &  \multicolumn{2}{c}{WoZ} \\
Model & Joint goal & Turn request\\
\toprule
 \small Delex. Model~\citep{Mrksic2017NeuralBT} & 70.8\% & 87.1\% \\
 \small Delex. + Semantic Dictionary~\citep{Mrksic2017NeuralBT} & 83.7\% & 87.6\% \\
 \small Neural Belief Tracker-DNN~\citep{Mrksic2017NeuralBT} & 84.4\% & 91.2\% \\
\small Neural Belief Tracker-CNN~\citep{Wen2017ANE} & 84.2\% & 91.6\% \\
\small GLAD~\citep{Zhong2018GlobalLocallySD} & 88.1$\pm$0.4\% &  97.1$\pm$0.2\% \\
\small GCE (Ours) & \textbf{88.51\%} & \textbf{97.38\%} \\
\bottomrule
\end{tabular}
\end{table}

\renewcommand{\tabcolsep}{0.3cm}
 \begin{table}
\caption{Time complexity for each batch of turn, and train and test epoch on WoZ dataset. Each batch contains 50 turns. All numbers are in second.}
\label{tab:time}
\centering
\begin{tabular}{lcccc}
\toprule
 &  \multicolumn{2}{c}{Train (sec.)} & \multicolumn{2}{c}{Test (sec.)}\\
Model & Turn & Total & Turn & Total \\
\toprule
\small GLAD~\citep{Zhong2018GlobalLocallySD} &  1.78 & 89 & 2.32 & 76 \\
\small GCE (Ours) & \textbf{1.16} & \textbf{60} &  \textbf{1.92} & \textbf{63} \\
\bottomrule
\end{tabular}
\end{table}

\begin{table}[t!]
\caption{Performance on Multi-Domain dataset, Multi-WoZ~\citep{Budzianowski2018MultiWOZA}. 
}
\label{tab:mwoz}
\centering
\begin{tabular}{lccc}
\toprule
& &  \multicolumn{2}{c}{Multi-WoZ} \\
Model & split & Turn inform & Joint goal \\
\toprule
\multirow{2}{*}{\small GLAD~\citep{Zhong2018GlobalLocallySD}} & Dev & 66.91\% & 34.83\% \\
& \small Test & 66.89\% & 35.57\% \\
\midrule
\multirow{2}{*}{\small GCE (Ours)} &  Dev & \textbf{67.78}\% & \textbf{37.42}\% \\
& \small Test & \textbf{67.88}\% & \textbf{35.58}\% \\
\bottomrule
\end{tabular}
\end{table}



\section{Conclusion}
In this paper, we proposed a neural model for dialogue state traking. Based on globally conditioning the encoder model on slot types (GCE), slot-conditioned representations are computed for user utterance and previous system actions, which are used to compute the mentioned slot value. By relaxing GLAD model from slot-specific recurrent networks and self-attentions, our model achieved lower computational complexity with better accuracy. We also showed that GCE model is generalizable to multi-domain dialogue state tracking, by evaluation on Multi-WoZ dataset. 

\bibliography{references}
\bibliographystyle{abbrvnat}

\end{document}